\documentclass{article}
\usepackage{spconf,amsmath,graphicx,hyperref}
\usepackage{booktabs}
\usepackage{amssymb}
\newcommand{\R}{\mathbb{R}}

\newcommand{\method}{\textsc{ALOE}}

\title{ALOE: SEMANTICALLY ADDRESSED LOW-RANK OPERATORS FOR KNOWLEDGE EDITING}
\name{Zeyan Li$^1$, Hu Xu$^1$, Jianfeng Xu$^1$\sthanks{Corresponding author.}}
\address{$^1$ Shanghai Jiao Tong University}

\begin{document}
\ninept
\maketitle

\begin{abstract}
Knowledge editing changes what a model knows by modifying parameters so that a requested fact updates while unrelated behavior is preserved. This is usually treated as a write problem, but editing also involves an address problem: deciding which hidden states should receive the new residual. An update that activates too narrowly memorizes one prompt, while one that activates too broadly disrupts neighboring knowledge. Parametric editors encode this scope implicitly, whereas memory-based editors make the selection explicit but keep it outside the edited model. We propose ALOE (Addressed Low-rank Operator for Editing), which learns semantic addresses from paraphrases and hard same-subject negatives, aligns them with autoregressive hidden states through rollout refinement and gate calibration, and embeds the resulting gated low-rank operator within one MLP layer, so that the deployed model runs in a single forward pass with no external retriever or auxiliary router. Evaluated on CounterFact, ZSRE, and KnowEdit across three 7--8B model families, ALOE achieves efficacy between 0.955 and 0.999 and locality between 0.981 and 1.000; mechanistic analyses confirm that the learned geometry separates competing edits and that calibration suppresses out-of-scope activation. The remaining errors concentrate in paraphrase coverage and write fitting.
\end{abstract}


\section{Introduction}
Language models store factual associations in their parameters, including feed-forward layers and individual neurons. Knowledge editing aims to revise such associations on request: the edited model should produce the new fact on the original request (efficacy), transfer the change to equivalent wordings of that request (generalization), and leave unrelated behavior untouched (locality). All three criteria depend on one choice that is rarely made explicit: where in activation space the update is allowed to act.

\begin{figure}[t]
  \centering
  \includegraphics[width=\columnwidth]{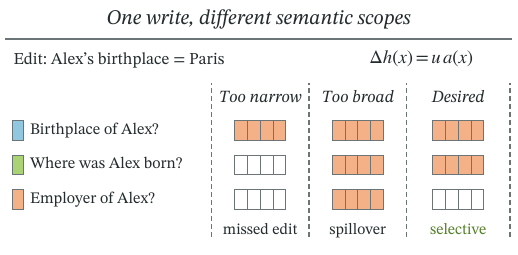}
  \caption{The address problem. The same write can miss paraphrases or
  affect neighboring facts. Filled cells denote an active write; empty cells
  denote no update (schematic).}
  \label{fig:address-problem}
\end{figure}

Editing therefore involves two distinct problems. The first is a \emph{write problem}: specifying the residual that encodes the new fact. The second is an \emph{address problem}: deciding which hidden states should receive that residual. A birthplace edit, for example, should cover rewordings of the same question without changing the model's answer about the person's employer, and at scale, different subjects that share a relation must receive distinct writes. An update whose address is too narrow memorizes the training prompt and misses its rewordings; one whose address is too broad activates on inputs it should leave alone (Fig.~\ref{fig:address-problem}). These boundary failures are documented in current editors: an edit perturbs other facts about the same subject \cite{ma-etal-2024-neighboring,duan-etal-2025-related}, damages knowledge beyond its intended scope \cite{wang2025missing}, and fails to propagate to facts implied by the edited one \cite{cohen-etal-2024-ripple}.

Existing editors handle the address problem in one of two ways. ROME and MEMIT encode facts through low-rank MLP updates, while PMET refines the participating states \cite{meng2022rome,meng2023memit,li-etal-2024-pmet}. In such an update $\Delta W=UV^\top$, which acts on the layer input $h(x)$ of a prompt $x$, the term $V^\top h(x)$ already serves as an address for each write column, but its factors are never trained to separate paraphrases from semantic neighbors, and causal localization need not identify the most editable layer \cite{hase-etal-2023-does}. SERAC, GRACE, and WISE instead make the selection explicit through a classifier, a codebook, or a side memory \cite{mitchell-etal-2022-memory,hartvigsen2023grace,wang2024wise}, but their deployed systems perform this selection outside the edited weights.

To obtain selection that is both explicitly learned and executed inside the edited model, we propose \method{} (Addressed Low-rank Operator for Editing). During construction, asymmetric query/key maps learn a scope representation for each edit from three kinds of examples: paraphrases of the request, locality inputs that should remain unchanged, and same-subject prompts whose relation differs from the edited one, which serve as hard negatives. The resulting codes initialize address vectors inside one MLP layer, and these addresses are refined and calibrated on autoregressive rollouts so that they respond to the states the model produces at generation time. A joint solve then fits the write directions under the calibrated gates. At inference, the edited model needs no external component, because the address is a trained part of the MLP itself. Our contributions are as follows. First, we formulate knowledge editing as a coupled address--write problem and show empirically that the two components fail independently, so each can be measured and diagnosed on its own. Second, we realize the address as an intrinsic gated low-rank MLP operator: it learns semantic edit boundaries at construction time and executes them inside the edited layer at inference. Third, on CounterFact, ZSRE, and KnowEdit across three 7--8B model families, \method{} attains efficacy between 0.955 and 0.999 and locality between 0.981 and 1.000 on edit streams of 839 to 1,301 facts.

\begin{figure*}[t]
  \centering
  % Exported from aloe-architecture-semantic.pptx (editable architecture master).
  \includegraphics[width=\textwidth]{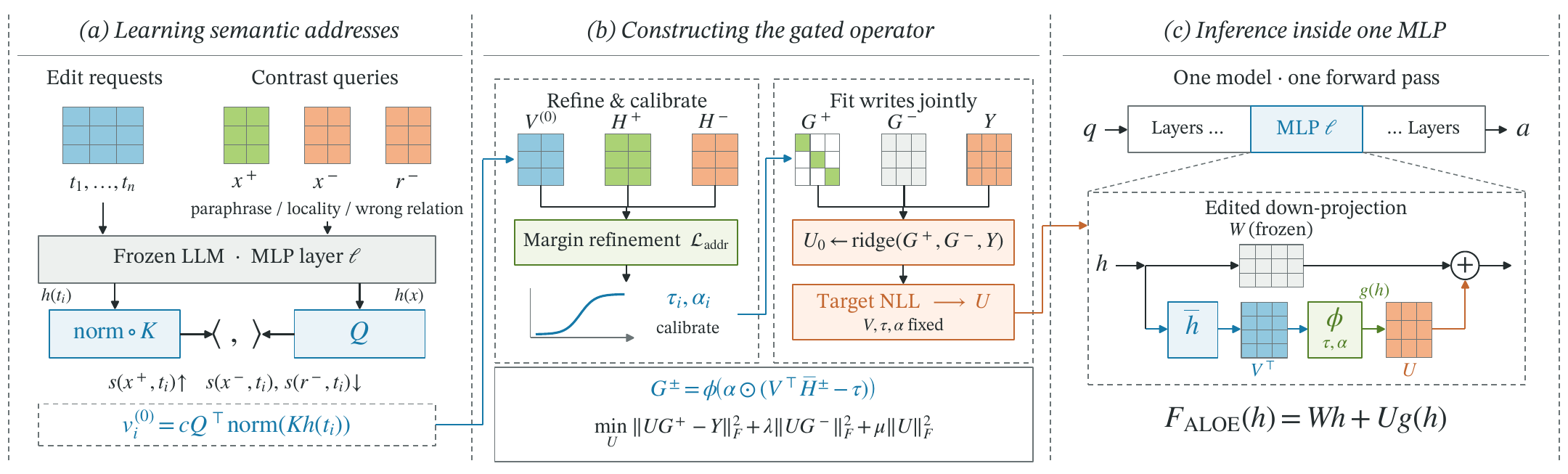}
  \caption{ALOE learns semantic addresses, refines and calibrates continuous
  edit gates, and inserts the resulting low-dimensional residual into one MLP
  projection.  The edited LLM executes the operator in a standard forward pass.}
  \label{fig:operator}
\end{figure*}

\section{Related work}

\subsection{Parametric knowledge editing}

Parametric editors differ mainly in how they construct and constrain weight updates. KnowledgeEditor and MEND learn to transform edit gradients into parameter changes, and MALMEN extends learned editing to large batches \cite{de-cao-etal-2021-editing,mitchell2022mend,tan-etal-2024-malmen}. ROME instead writes a factual association directly into an MLP through a rank-one update, and MEMIT extends this construction to many edits at once \cite{meng2022rome,meng2023memit}; carefully configured fine-tuning remains a useful reference. More recent methods such as AlphaEdit constrain updates to preservation or orthogonal subspaces to reduce interference with existing knowledge \cite{fang2025alphaedit,xu-etal-2025-knowledge}.

These methods decide \emph{how} and \emph{where} new knowledge is written into shared parameters. The scope of the resulting edit---which inputs should activate it---is induced implicitly by the interaction between the update and the model's hidden states.

\subsection{Selective access and edit scope}

A second line of work makes access to edited knowledge explicit. SERAC retrieves a counterfactual model, GRACE stores edits in discrete key--value adaptors, and WISE routes inputs among memories \cite{mitchell-etal-2022-memory,hartvigsen2023grace,wang2024wise}. In-context editing avoids parameter modification altogether by placing updated facts in demonstrations \cite{zheng-etal-2023-can}. Locate-then-edit and associated-knowledge methods improve the placement or propagation of edits. In all of these systems, the selection mechanism lives outside the edited weights---a retriever, a codebook, a router, or the prompt---or targets where an edit should be placed rather than when it should fire. ALOE learns semantic addresses that gate low-rank writes directly inside an edited MLP, so selective access becomes part of the deployed model itself.

The scope of an edit is also central to how editing is evaluated. CounterFact+ strengthens tests of specificity \cite{hoelscher-obermaier-etal-2023-detecting}, and RippleEdits and ReCoE measure whether an edit propagates to related knowledge \cite{cohen-etal-2024-ripple,hua-etal-2024-propagation}. Sequential-edit studies show that poorly controlled updates cause forgetting and degrade general abilities. These evaluation results point to a shared requirement: an edit should activate for equivalent or relevant queries and remain inactive for nearby but out-of-scope knowledge. We formulate this boundary explicitly as the \emph{address} of an edit and study it separately from the write itself.

\section{Method}
\subsection{An addressed residual operator}
Figure~\ref{fig:operator} illustrates the construction pipeline. Let $f_\theta$ be a frozen transformer, and let $\mathcal E=\{(t_i,y_i)\}_{i=1}^{n}$ denote the requested edits, where $t_i$ is the request text and $y_i$ is its target. We augment a single MLP down-projection $W\in\R^{d_{\mathrm{out}}\times d}$, whose input state is $h\in\R^d$. ALOE gives each edit two learned vectors: an \emph{address} that decides which hidden states the edit applies to, and a \emph{write direction} that carries the new fact. The address gates the write direction, so the edit only enters the output where its address activates:
\begin{align}
 \bar h &= h/\max(\|h\|_2,\epsilon), \nonumber\\
 g(h) &= \phi\!\left(\boldsymbol\alpha\odot
      (V^\top\bar h-\boldsymbol\tau)\right), \label{eq:gate}\\
 F_{\mathrm{ALOE}}(h) &= Wh+Ug(h). \label{eq:update}
\end{align}
Here $V=[v_1,\ldots,v_n]\in\R^{d\times n}$ contains the semantic addresses, $U=[u_1,\ldots,u_n]\in\R^{d_{\mathrm{out}}\times n}$ contains the write directions, and $\boldsymbol\tau,\boldsymbol\alpha\in\R^n$ are per-edit thresholds and temperatures. Normalizing the input state in $\bar h$ makes each gate depend on the direction of the hidden state rather than its magnitude. The dead-zone sigmoid $\phi(z)=\max(\sigma(z)-\epsilon_g,0)/(1-\epsilon_g)$ with $\epsilon_g=10^{-3}$ drives the gates of inactive addresses to exactly zero while retaining continuous activation above threshold, so an edit whose address does not match has no effect on the output. The operator is therefore state-dependent: the hidden state at each token determines which write directions enter the output, in contrast to the fixed linear update $W+UV^\top$ of standard low-rank editing.

\subsection{Learning the semantic boundary}
The first construction stage learns where each residual should act. For edit $i$, the full request serves as the canonical key, so that the address retains the identity of both the subject and the relation. Each training item contains a paraphrase $x_i^+$ of the request, a locality input $x_i^-$ that should not activate the edit, and same-subject prompts $r_{ik}^-$ whose predicates differ from $t_i$; because these prompts share the subject with the edit, they are harder negatives than random prompts.

We learn two distinct linear maps $Q,K\in\R^{p\times d}$ and a positive scale $c$. With $k_i=\operatorname{norm}(Kh(t_i))$ as the key of edit $i$, the construction-time score between a query $x$ and edit $i$ is
\begin{equation}
    s(x,t_i)=c\langle Qh(x),k_i\rangle .
    \label{eq:score}
\end{equation}
A high score means the query falls inside the edit's scope. The training loss ranks $s(x_i^+,t_i)$ above the scores of wrong-relation queries and keys, separates different requests within a minibatch, and enforces the locality margin $s(x_i^-,t_i)+m<s(x_i^+,t_i)$. Orthogonality regularization discourages the keys of different edits from collapsing onto each other, and relation-balanced minibatches prevent frequent predicates from dominating the loss. Labels and contrast sets are used only during construction; once the addresses are built, inference relies on the resulting address parameters alone.

\subsection{From construction to generation}
The metric learned above separates construction prompts, but the deployed address must fire on hidden states encountered during generation, which need not coincide with the states seen at construction time: once the model begins producing its own tokens, its hidden states drift away from those measured on pre-written prompts. To bridge this shift, we initialize the address of edit $i$ as
\begin{equation}
 v_i^{(0)}=cQ^\top\operatorname{norm}\!\left(Kh(t_i)\right), \qquad c>0,
 \label{eq:key}
\end{equation}
where $\operatorname{norm}(z)=z/\|z\|_2$. Transposition preserves the ordering induced by the learned metric, and the normalization in Eq.~\eqref{eq:gate} removes the magnitude of the hidden state. We then collect normalized anchor states $H^+$ and negative states $H^-$ from left-padded autoregressive rollouts, in which the model generates continuations of each request and we record the states it passes through, and refine $V^{(0)}$ on these generation-time states by
\begin{equation}
 \mathcal L_{\mathrm{addr}}=\frac1n\sum_i
 \left[m_i-\min_{h\in\mathcal P_i}v_i^\top h
 +\max_{h\in\mathcal N_i}v_i^\top h\right]_+^2,
 \label{eq:address-refine}
\end{equation}
for up to 3,000 AdamW steps at learning rate 0.01, preserving the norms of the addresses. This loss pushes each address toward its worst-scoring anchor state and away from its worst-scoring negative state, so that even the least typical valid phrasing still opens the gate. Edits with identical input--target pairs share one address support, while conflicting targets remain separate constraints. Calibration then converts the refined scores into gates. For a separable slot $i$, it sets $\tau_i$ and $\alpha_i$ so that the worst negative falls below the dead zone and the worst positive maps to 0.9; for a slot whose states cannot be separated, a finite-temperature midpoint initialized at 8 minimizes a balanced worst-case classification loss. After refinement, the metric-learning components are discarded: $Q$, $K$, the labels, and the prompts are no longer needed, and only $V$, $\boldsymbol\tau$, and $\boldsymbol\alpha$ remain in the edited model.

\subsection{Fitting writes jointly}
Once the address support is fixed, the remaining task is to fit write directions that are compatible with the gates. Let $Y$ contain target residuals obtained by independently optimizing each edit for 25 steps, compressing its down-projection change to rank 16, and evaluating the compressed change on its anchor; column $i$ of $Y$ is thus the output change that edit $i$ should produce when its gate is fully open. With $G^+=g(H^+)$ and $G^-=g(H^-)$ collecting the gate activations of anchor and negative states, a preservation-regularized ridge solve initializes all writes jointly:
\begin{equation}
 U_0=YG^{+\top}\!\left(G^+G^{+\top}+\lambda G^-G^{-\top}+\mu I\right)^{-1}.
 \label{eq:solve}
\end{equation}
The gates of different edits can overlap, so one edit's write direction can leak through another edit's gate; solving for all writes in one system accounts for this cross-activation, and the $G^-G^{-\top}$ term keeps the writes small on states where no edit should fire. With $V$, $\boldsymbol\tau$, and $\boldsymbol\alpha$ frozen, we then optimize $U$ from $U_0$ for up to 100 AdamW steps using the negative log-likelihood of the autoregressive targets. At deployment, the model computes all gates densely and adds $Ug(h)$ to $Wh$, which requires $O(n(d+d_{\mathrm{out}}))$ storage and compute linear in the number of edits.

\begin{table*}[t]
\centering
\caption{Efficacy (Eff.), generalization (Gen.), and locality (Loc.) across
base models and benchmarks. ALOE fits one operator to the complete benchmark
stream (839 CounterFact, 1,266 KnowEdit, and 1,301 ZSRE edits); all results
are means over seeds 0, 42, and 99. Best and second-best values are
\textbf{bolded} and \underline{underlined}, respectively.}
\label{tab:baseline-model-matrix}

\fontsize{9}{9.1}\selectfont
\renewcommand{\arraystretch}{0.9}
\setlength{\tabcolsep}{0.4pt}

\begin{tabular*}{\textwidth}{
@{\extracolsep{\fill}}
l*{18}{c}
@{}
}
\toprule
& \multicolumn{3}{c}{\textbf{ALOE}}
& \multicolumn{3}{c}{ROME (2022)}
& \multicolumn{3}{c}{MEMIT (2023)}
& \multicolumn{3}{c}{MEND (2022)}
& \multicolumn{3}{c}{GRACE (2023)}
& \multicolumn{3}{c}{AlphaEdit (2025)} \\
\cmidrule(lr){2-4}
\cmidrule(lr){5-7}
\cmidrule(lr){8-10}
\cmidrule(lr){11-13}
\cmidrule(lr){14-16}
\cmidrule(lr){17-19}

Benchmark
& Eff. & Gen. & Loc.
& Eff. & Gen. & Loc.
& Eff. & Gen. & Loc.
& Eff. & Gen. & Loc.
& Eff. & Gen. & Loc.
& Eff. & Gen. & Loc. \\
\midrule

\multicolumn{19}{l}{\textbf{Llama-2-7B}} \\[-1pt]

CounterFact
& \textbf{0.956}
& \textbf{0.289}
& \textbf{0.982}
& 0.093 & 0.083 & 0.014
& 0.056 & \underline{0.188} & 0.030
& 0.011 & 0.012 & 0.005
& \underline{0.949} & 0.003 & \underline{0.972}
& 0.821 & 0.129 & 0.838 \\

KnowEdit
& \textbf{0.999}
& \underline{0.472}
& \textbf{1.000}
& 0.121 & 0.106 & 0.028
& 0.162 & 0.150 & \underline{0.195}
& 0.002 & 0.002 & 0.002
& \underline{0.980} & 0.086 & \textbf{1.000}
& 0.969 & \textbf{0.801} & \textbf{1.000} \\

ZSRE
& \textbf{0.997}
& \underline{0.464}
& \textbf{1.000}
& 0.204 & 0.183 & 0.019
& 0.144 & 0.138 & \underline{0.137}
& 0.003 & 0.003 & 0.003
& 0.974 & 0.009 & \textbf{1.000}
& \underline{0.977} & \textbf{0.891} & \textbf{1.000} \\

\midrule

\multicolumn{19}{l}{\textbf{Llama-3.1-8B-Instruct}} \\[-1pt]

CounterFact
& \underline{0.959}
& \underline{0.281}
& \textbf{0.981}
& 0.096 & 0.086 & 0.003
& 0.031 & 0.006 & 0.341
& 0.010 & 0.010 & 0.005
& 0.423 & 0.002 & \underline{0.971}
& \textbf{0.982} & \textbf{0.813} & 0.325 \\

KnowEdit
& \textbf{0.999}
& \underline{0.459}
& \textbf{1.000}
& 0.141 & 0.108 & 0.010
& 0.079 & 0.073 & 0.028
& 0.026 & 0.026 & 0.008
& 0.350 & 0.018 & \textbf{1.000}
& \underline{0.996} & \textbf{0.798} & \underline{0.578} \\

ZSRE
& \textbf{0.995}
& \underline{0.407}
& \textbf{1.000}
& 0.162 & 0.138 & 0.020
& 0.101 & 0.097 & 0.151
& 0.007 & 0.007 & 0.006
& 0.343 & 0.019 & \textbf{1.000}
& \underline{0.994} & \textbf{0.901} & \underline{0.527} \\

\midrule

\multicolumn{19}{l}{\textbf{Qwen3-8B}} \\[-1pt]

CounterFact
& \underline{0.955}
& 0.217
& \textbf{0.981}
& \textbf{0.979} & \textbf{0.571} & 0.091
& 0.859 & \underline{0.542} & 0.276
& 0.012 & 0.013 & 0.007
& 0.414 & 0.001 & \underline{0.971}
& 0.839 & 0.533 & 0.305 \\

KnowEdit
& \textbf{0.998}
& 0.359
& \textbf{1.000}
& \underline{0.986} & \underline{0.625} & 0.194
& 0.946 & \textbf{0.636} & \underline{0.450}
& 0.005 & 0.004 & 0.002
& 0.349 & 0.008 & \textbf{1.000}
& 0.817 & 0.554 & 0.432 \\

ZSRE
& \textbf{0.994}
& 0.376
& \textbf{1.000}
& \underline{0.981} & \textbf{0.898} & 0.202
& 0.845 & 0.751 & 0.447
& 0.006 & 0.005 & 0.005
& 0.343 & 0.011 & \textbf{1.000}
& 0.914 & \underline{0.796} & \underline{0.483} \\

\bottomrule
\end{tabular*}
\end{table*}

\section{Experiments}

\subsection{Setup}
We evaluate end-to-end editing on CounterFact \cite{meng2022rome}, ZSRE \cite{levy-etal-2017-zero}, and KnowEdit \cite{zhang-etal-2024-comprehensive}, covering factual replacement, question-answer rephrasing, and diverse knowledge domains. The main comparison uses Llama-2-7B, Llama-3.1-8B-Instruct, and Qwen3-8B \cite{touvron-etal-2023-llama,grattafiori-etal-2024-llama,qwen-team-2025-qwen3}, and the address-transfer study adds Qwen2.5-7B-Instruct \cite{qwen-team-2024-qwen25}. All models are run with seeds 0, 42, and 99, and every result reported in this section is the mean over the three runs. The comparison includes five established knowledge editors: ROME \cite{meng2022rome}, MEMIT \cite{meng2023memit}, MEND \cite{mitchell2022mend}, GRACE \cite{hartvigsen2023grace}, and AlphaEdit \cite{fang2025alphaedit}. We report the standard dimensions separately: \emph{efficacy} is exact match on the edited request, \emph{generalization} is exact match on rephrases, and \emph{locality} is normalized exact-generation stability on out-of-scope prompts relative to the unedited model. Baselines use their published configurations through EasyEdit where supported \cite{wang-etal-2024-easyedit}.

\subsection{End-to-end editing}
Table~\ref{tab:baseline-model-matrix} reports the three metrics separately for each model--benchmark pair. The baselines split into two failure patterns. On Llama-2-7B and Llama-3.1-8B-Instruct, the locate-and-edit and meta-learning methods collapse under edit streams of this length: ROME, MEMIT, and MEND lose almost all efficacy on CounterFact, and GRACE retains efficacy only on Llama-2 while activating on almost no rephrase anywhere. On Qwen3-8B these methods survive, but the survivors pay for generalization with locality: ROME's rephrase accuracy comes with locality at or below 0.20, and MEMIT and AlphaEdit show the same exchange at milder levels. AlphaEdit is the strongest baseline overall, yet its profile is uneven across model families---competitive on Llama-2, it keeps high generalization on Llama-3 only while locality falls to 0.33--0.58, and on Qwen3 both its efficacy and its locality drop well below ALOE's. ALOE is the only method whose efficacy stays between 0.955 and 0.999 and whose locality stays between 0.981 and 1.000 in all nine model--benchmark cells, so its advantage is consistency: no failure cell, on any model family, under streams of up to 1,301 edits. The cost is equally visible. Generalization ranges from 0.217 to 0.472, below the best baseline cells on each model.

\begin{figure}[!t]
  \centering
  \includegraphics[width=\columnwidth]{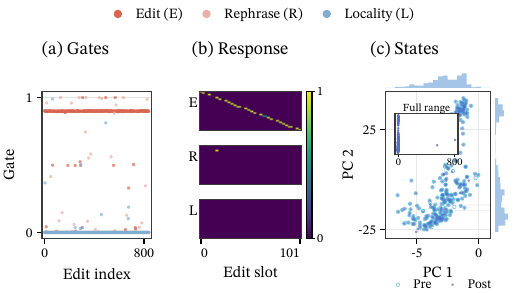}
  \caption{Runtime behavior after 839 CounterFact edits on Llama-3.1-8B-Instruct. (a) Matched-slot gates for edits and
  rephrases, and the maximum gate for locality queries. (b) Responses of the
  first 32 sampled queries to their associated slots. (c) Paired final-block
  locality states under shared PCA, with empirical marginals and a full-range
  inset. PC1/PC2 explain 45.1\%/10.7\%; mean relative state drift is 11.54\%
  and next-token agreement is 96.5\%.}
  \label{fig:mechanism-evidence}
\end{figure}

\subsection{Mechanism analysis}
We test the address mechanism on its own, using disjoint CounterFact train, calibration, and development partitions organized by predicate, with development subjects and exact surfaces excluded from training; the development set contains 632 examples from 34 predicates, each query paired with 16 fixed same-subject wrong-relation keys. We measure scope AUC, calibration-threshold recall, 16-way same-subject top-1, and transferred-threshold negative FPR against equal-dimensional raw activation controls, with all runs at seeds 0, 42, and 99. The learned address beats the raw controls by a wide margin: AUC 0.9976 against 0.6365, recall 0.9910 against 0.1804, same-subject top-1 0.8576 against 0.3117, and FPR 0.0418. Standard deviations across seeds stay below 0.006 on every metric, so the selectivity comes from the learned geometry rather than from the hidden states themselves. The choice of key matters just as much: an entity-neutral key collapses 632 edits onto 179 unique addresses, while the full request keeps all 632 distinct and raises synthetic-write top-1 from 0.698 to 0.998. The same construction objective transfers without retuning to the other three model families, which reach AUC 0.997--0.998 and FPR 0.036--0.042; their slightly lower same-subject top-1 (0.830--0.848) points to architecture-dependent separation margins.

The deployed checkpoint tells the same story from the runtime side and locates the remaining failures. On 256 fixed CounterFact cases drawn from the 839-edit model (Fig.~\ref{fig:mechanism-evidence}), the matched slot dominates 98.4\% of original requests, and locality states drift by 11.54\% on average with a median of zero, consistent with the near-perfect locality scores; the visible weakness is on rephrases, whose mean gate activation is 0.057 against 0.860 for original requests, so most rephrasings never open the gate. Controlled interventions on 256 CounterFact edits, removing one construction stage at a time with the data, layer, and write budget fixed, assign this gap to specific stages: rollout refinement contributes 55.0 points of efficacy, confirming that the construction metric must be aligned with generation-time states; calibration contributes 24.8 points of efficacy and 96.5 points of locality while slightly reducing generalization, because thresholds suppress false activation but cannot create paraphrase support that the address lacks; the preservation term changes every measure by at most half a point. Paraphrase coverage is therefore an address problem, visible in the gate traces, while the residual gap between an open gate and a correct generation is a write-fitting problem, visible in the transfer study where address quality holds but end-to-end accuracy does not.

\section{Conclusion}
We formulated knowledge editing as a coupled address--write problem and proposed ALOE, which makes the address intrinsic to an edited MLP. An asymmetric metric learns the scope of each edit from paraphrases and same-subject hard negatives, rollout refinement and gate calibration align this scope with generation-time hidden states, and a joint solve fits the write directions under the calibrated gates. On CounterFact, ZSRE, and KnowEdit across three 7--8B model families, ALOE attains efficacy between 0.955 and 0.999 and locality between 0.981 and 1.000 on streams of up to 1,301 edits. Mechanism analyses show that the learned addresses separate in-scope states from competing ones well beyond what raw hidden states provide, and that the same geometry transfers to an untuned model family. Controlled ablations further show that rollout refinement drives most of the efficacy improvement, while gate calibration drives most of the locality improvement. The remaining errors follow the same decomposition: missed paraphrases are address failures, while an open gate followed by a wrong generation is a write failure. Generalization remains between 0.217 and 0.472 because calibrated gates rarely open for rephrasings far from the training paraphrases, which is the clearest limitation of the current system. Our evaluation is limited to three English factual benchmarks, 7--8B models, and a single edited layer, with storage growing linearly in the number of edits.

\vfill\pagebreak

\bibliographystyle{IEEEbib}
\bibliography{references}

\end{document}